\documentclass{article}

\usepackage[preprint]{neurips_2021}

\usepackage{soul}
\usepackage{url}
\usepackage[hidelinks]{hyperref}
\usepackage{graphicx}
\usepackage{amsmath}
\usepackage{amsthm}
\usepackage{booktabs}
\usepackage{algorithm}
\usepackage{algorithmic}
\usepackage{hyperref}       
\usepackage{url}            
\usepackage{booktabs}       
\usepackage{amsfonts}       
\usepackage{nicefrac}       
\usepackage{microtype}      

\usepackage{multirow}
\usepackage{amssymb}
\usepackage{gensymb}
\usepackage{amsmath}
\usepackage{mathtools}
\usepackage{amsthm}
\usepackage{calrsfs}
\DeclareMathAlphabet{\pazocal}{OMS}{zplm}{m}{n}
\usepackage{makecell}
\usepackage{boldline}
\usepackage{xcolor}
\usepackage{color}
\usepackage{booktabs}

\newcommand*{\pd}[3][]{\ensuremath{\frac{\partial^{#1} #2}{\partial #3}}}
\newcommand{\ads}{\text{autonomous driving system }}

\title{Finding Adversarial Examples for Simulated Autonomous Driving with Fast and Differentiable Image Compositing}

\author{
    Jinghan Yang, Adith Boloor, Ayan Chakrabarti, Xuan Zhang, Yevgeniy Vorobeychik\\
    Washington University in St. Louis\\
    \{jinghan.yang, adith, ayan, xuan.zhang, yvorobeychik\}@wustl.edu
}

\begin{document}

\maketitle

\begin{abstract}
There is considerable evidence that deep neural networks are vulnerable to adversarial perturbations applied directly to their digital inputs. However, it remains an open question whether this translates to vulnerabilities in real systems.
For example, an attack on self-driving cars would in practice entail modifying the driving environment, which then impacts the video inputs to the car's controller, thereby indirectly leading to incorrect driving decisions.
Such attacks require accounting for system dynamics and tracking viewpoint changes. 
We propose a scalable approach for finding adversarial modifications of a simulated autonomous driving environment using a differentiable approximation for the mapping from environmental modifications (rectangles on the road)
to the corresponding video inputs to the controller neural network. 
Given the 
parameters of the rectangles, our proposed differentiable mapping composites them onto pre-recorded video streams of the original environment, accounting for geometric and color variations. 
Moreover, we propose a multiple trajectory sampling approach that enables our attacks to be robust to a car's self-correcting behavior.
When combined with a neural network-based controller, our approach allows the design of adversarial modifications through end-to-end gradient-based optimization. 
Using the Carla autonomous driving simulator, we show that our approach is significantly more scalable and far more effective at identifying autonomous vehicle vulnerabilities in simulation experiments than a state-of-the-art approach based on Bayesian Optimization.
\end{abstract}

\section{Introduction}

Computer vision has made revolutionary advances in recent years by leveraging a combination of deep neural network architectures with abundant high-quality perceptual data.
One of the transformative applications of computational perception is autonomous driving, with autonomous cars and trucks already being evaluated for use in geofenced settings, and partial autonomy, such as highway assistance, leveraging state-of-the-art perception embedded in vehicles available to consumers.
However, a history of tragic crashes involving autonomous driving, most notably Tesla~\citep{Thorbecke20} and Uber~\citep{Hawkins19} reveals that modern perceptual architectures still have some limitations even in non-adversarial driving environments.
Even more concerning is the increasing abundance of evidence that state-of-the-art deep neural networks used in perception tasks are vulnerable to \emph{adversarial perturbations}, or imperceptible noise added to an input image and designed to cause misclassification~\citep{goodfellow2014explaining,yuan2019adversarial}.
Furthermore, several lines of work consider specifically \emph{physical adversarial examples} which modify the \emph{scene} being captured by a camera, rather than the image~\citep{Kurakin_2016_CoRR,Eykholt_2018_CVPR,Sitawarin_2018_CoRR,dutta2018security,Duan_2020_CVPR}.

Despite this body of evidence demonstrating vulnerabilities in deep neural network perceptual architectures, it is not evident that they are consequential in realistic  autonomous driving, even if primarily using cameras for perception.
First, most such attacks involve independent perturbations to input images.
Autonomous driving is a dynamical system, so that a fixed adversarial perturbation to a scene is perceived through a series of highly interdependent perspectives.
Second, even if we succeed in causing the control outputs of self-driving cars to deviate from normal, the vehicle will now perceive a sequence of frames that is \emph{different from those encountered on its normal path}, and typically deploy self-correcting behavior in response.
For example, if the vehicle is driving straight and then begins swerving towards the opposite lane, its own perception will inform the control that it's going in the wrong direction, and the controller will steer it back on course.

\begin{figure*}[t]
  \centering
  \includegraphics[width=\textwidth]{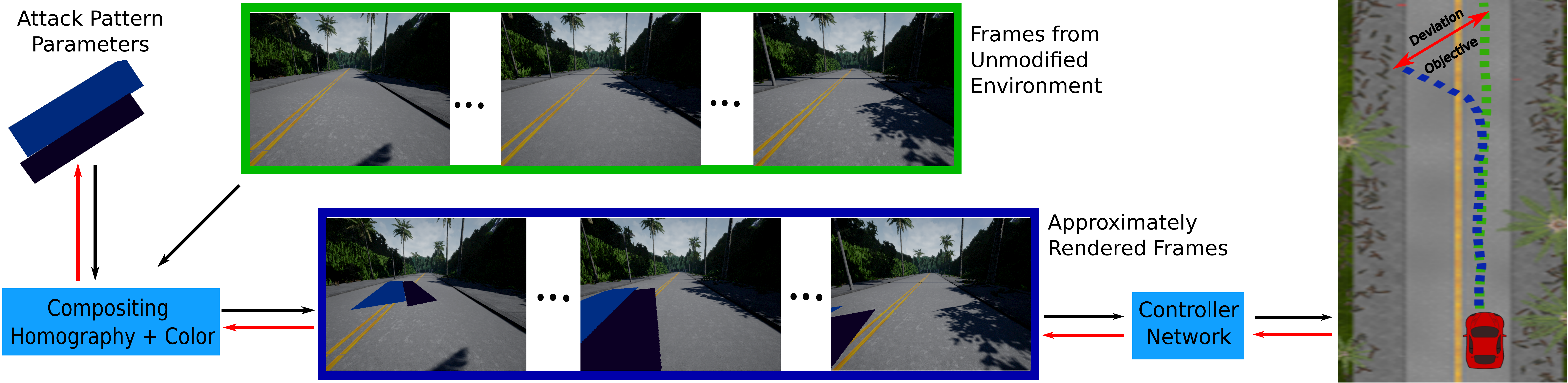}
  \caption{Overview. We collect and calibrate frames from the unmodified environment (shown in the green box), and given a choice of attack pattern parameters, composite the pattern to create approximate renderings of frames corresponding to placing the pattern in the environment. Our composition function is differentiable with respect to the attack pattern parameters, and we are thus able to use end-to-end gradient-based optimization when attacking a differentiable control network, to cause the network to output incorrect controls that cause the vehicle to deviate from its intended trajectory (from the green to the blue trajectory, as shown in the right column), and crash.
  } 
\label{fig:crash}
\end{figure*}
To address these limitations, \citet{Boloor_2020_JSA} recently proposed Bayesian Optimization (BO)
as a way to design simple \emph{physically realizable} (that is, easy to implement in a physical scene) adversarial examples 
in Carla autonomous driving simulations~\citep{Dosovitskiy17} against end-to-end autonomous driving architectures.
As simulations are a critical part of autonomous vehicle development and testing~\citep{Dosovitskiy17,Waymo20}, this was an important advance in enabling scalable identification of practically consequential vulnerabilities in autonomous vehicle architectures.
The key challenge with this approach, however, is that attack design must execute actual simulation experiments 
for a large number of iterations (1000 in the work above), making it impractical for large-scale or physical driving evaluation.

We propose a highly scalable framework for designing physically realizable adversarial examples for adversarial testing of simulated end-to-end autonomous driving architectures.
Our framework is illustrated in Figure~\ref{fig:crash}, and develops a differentiable pipeline for digitally approximating driving scenarios.
The proposed approximation makes use of image compositing, learning homography and color mappings from a birds-eye view of embedded adversarial examples to projections of these in images based on actual driving frames, and sampling sequences of actual frames with small perturbations to control to ensure adequate sampling of possible perspectives.
The entire process can then be fed into automatic differentiators to obtain adversarial examples that maximize a car's deviation from its normal sequence of controls (e.g., steering angle) for a target driving scenario.

We evaluate the proposed framework using Carla simulations in comparison with the state-of-the-art BO method.
Our experiments show that the resulting attacks are significantly stronger, with effects on induced deviations and road infractions often considerably outperforming BO, using an order of magnitude fewer simulation runs.
Furthermore, we show that our approach yields adversarial test instances that are robust to unforeseen variations in weather and visibility.




\noindent{\bf Related Work: }
Attacks on deep neural networks for computer vision tasks has been a subject of extensive prior research~\citep{goodfellow2014explaining,yuan2019adversarial,Modas_2020_IEEE,VorobeychikBook18}. 
The most common variation introduces imperceptible noise to pixels of an image in order to induce error in predictions, such as misclassification of the image or failure to detect an object in it.
A more recent line of research has investigated \emph{physical adversarial examples} \citep{Kurakin_2016_CoRR,Athalye_2017_CoRR,Eykholt_2018_CVPR,Sitawarin_2018_CoRR,dutta2018security,Duan_2020_CVPR}, where the explicit goal is to implement these in the physical scene, so that the images of the scene subsequently captured by the camera and fed into a deep neural network result in a prediction error.
In a related effort, \citet{Liu_2018_CoRR} developed a differentiable renderer that allows the attacker to devise higher-level perturbations of an image scene, such as geometry and lighting, through a differentiable renderer.
However, most of these approaches attack a fixed input scene, whereas autonomous driving is a complex dynamical system.
Several recent approaches investigate physical attacks on autonomous driving that attempt to account for the fact that a single object is modified and viewed through a series of frames~\citep{ackerman_2019,Kong_2019_CoRR,Boloor_2020_JSA}.
However, these either still consider digital attacks, albeit restricted to a small area (e.g., replacing a road sign with a noisy road sign) and do not consider a vehicles self-correcting behavior (for example,~\citet{Kong_2019_CoRR}), or rely on many expensive driving experiments in order to identify adversarial patterns~\citep{Boloor_2020_JSA}.
\section{Proposed Method}
\label{sec:Technique}

Autonomous driving systems are equipped with decision algorithms that produce control signals for a vehicle based on high-level instructions---such as a given route or destination---and inputs from cameras and other sensors that make continuous measurements of the vehicle's physical environment. We assume that the decision algorithm is in the form of a differentiable function---such as a neural network---that maps video frames from the camera, along with other inputs, to the control outputs. Given such a network or function, our goal is to determine if it is vulnerable to attack. Specifically, we seek to build a scalable and efficient method to find modifications that can be applied to a simulated autonomous driving environment with a sophisticated physics engine, and result in a stream of video frames which cause the control network to produce output signals that disrupt the vehicle's operation, moving it away from the expected ideal trajectory.

This task is challenging since the relationship between modifications to the simulated physical environment and the network's inputs is complex: the video frames correspond to images of the environment from a sequence of changing viewpoints, where the sequence itself depends on the network's control outputs. The precise effect of any given modification can be determined only by actually driving the vehicle in the modified simulated environment that uses
a physics engine. 
However, it is expensive to use such a simulator
when searching for the right modification, since the process is not differentiable with respect to parameters of the modification, and would require repeated trials with candidate modifications in every step of the search process.

Instead, we propose a fast approximation to produce video frames for a given environment given a candidate modification that is differentiable with respect to parameters of the modification. Our approach requires a small number of initial simulated calibration runs,
after which the search for optimal parameters can be carried out with end-to-end gradient-based optimization. Specifically, we consider the case when the modification takes the form of figures (such as rectangles) drawn on a restricted stretch of the road, and task the optimization with finding their optimal shape and color so as to maximize deviation from the controller's trajectory prior to modification. We now describe our model for the physical modification, our approximate mapping to create video frames for a given modification, and our optimization approach based on this mapping.

\subsection{Parameterized Scene Modifications}
\label{sec:physmap}

We assume modifications are in the form of a collection of $K$ figures (e.g., rectangles) that will be painted on a flat surface in the environment (e.g., road). 
Let $\Phi=\{x_k^S, x_k^C\}_{k=1}^K$ denote the parameters of this modification, with $x_k^S$ corresponding to the shape parameters, and $x_k^C$ the RGB color, of the $k^{th}$ figure. These parameters are defined with respect to co-ordinates in some canonical---say, top-down---view of the surface.

We let $M(n_c; x^S)$ denote a scalar-valued mask that represents whether a pixel at spatial location $n_c\in\mathbb{Z}^2$ in the canonical view is within a figure with shape parameters $x^S$. This function depends simply on the chosen geometry of the figures, and has value of $1$ for pixels within the figure, $0$ for those outside, and real values between $0$ and $1$ for those near the boundary (representing partial occupancy on a discrete pixel grid to prevent aliasing artifacts).

Since the spatial extents for different figures may overlap, we next account for occlusions by assuming that the lines will be painted in order. Accordingly, we define a series of visibility functions $V_k(n_c; \Phi)$, each representing the visibility of the $k^{th}$ figure at pixel $n_c$, after accounting for occlusions. We set the function for the last figure as $V_K(n_c; \Phi) = M(n_c; x_K^S)$, and for the other figures with $k < K$,
\begin{equation}
  V_k(n_c; \Phi) = M(n_c; x_k^S) \prod_{k'=k+1}^K \left(1-V_{k'}(n_c; \Phi)\right).
\end{equation}

\subsection{Approximate Frames via Compositing}
\label{sec:composite}

The next step in our pipeline deals with generating the video inputs that the controller network is expected to receive from a modified environment for given parameter values $\Phi$. These frames will represent views of the environment, including the surface with the painted figures, from a sequence of viewpoints as the car drives through the scene. Of course, the precise viewpoint sequence will depend on the trajectory of the car, which will depend on the control outputs from the network, that in turn depends on the frames. Instead of modeling the precise trajectory for every modification, we consider a small set of $T$ representative trajectories, corresponding to those that the vehicle will follow when driven with small perturbations around control outputs, when operating in the unmodified environment. 
One trajectory involves driving the car with the actual output control signals.
To generate others, we consider two variants: 1) adding random noise to control outputs (\emph{Random}), and 2) adding trajectories in pairs, one with random deviations to the left only, and the other only including random deviations to the right (termed \emph{Group}).
Given the fact that actual control is closed-loop, it is not evident that either variant of this simple approach would work; however, our experiments below (using $T=3$) show that both ideas are remarkably effective.

This gives $T$ sequences of video frames, one for each trajectory, where we assume each sequence contains $F$ frames. We let $\tilde{I}_{f}^t(n)$ denote the $f^{th}$ ``clean'' image in the $t^{th}$ sequence, representing a view of the environment without any modifications. Here, $n\in\mathbb{Z}^2$ indexes pixel location within each image, and the intensity vector $\tilde{I}_f^t(n)\in\mathbb{R}^3$ at each location corresponding to the recorded RGB values. These clean images can be obtained by driving the car---actually, or in simulation---in the original environment.

For each frame in each sequence, we also determine a spatial mapping $n_c = G_{f}^t(n)$ that maps pixel locations in the image to the canonical view. We model each $G_f^t(n)$ as a homography: the parameters of which can be determined by either using correspondences between each image and the canonical view of the surface---from calibration patterns rendered using the simulator, or from user input---or by calibrating the vehicle's camera and making accurate measurements of ego-motion when the vehicle is being driven. Additionally, we also determine color mapping parameters $C_f^t \in \mathbb{R}^{3\times 3}, b_f^t \in \mathbb{R}^3$ for each frame representing an approximate linear relationship between the RGB colors $x^C$ of the painted figures, and their colors as visible in each frame. These parameters are also determined through calibration.
Given this set of clean frames and the geometric and color mapping parameters, we generate corresponding frames with views of the modified environment simply as:
\begin{align}
  \label{eq:composite}
  \begin{split}
  I_f^t(n; \Phi)= \left(1 - \sum_{k=1}^K V_k(G_f^t(n); \Phi)\right) \tilde{I}_f^t(n) 
  + \sum_{k=1}^K V_k(G_f^t(n); \Phi)\left(C_f^t x_k^C + b_f^t\right).
  \end{split}
\end{align}

\subsection{Computing Adversarial Perturbations}
\label{sec:gopt}

Given the ``forward'' process of generating a set of frames for a given set of modification values $\Phi$, we finally describe our approach to finding the value of $\Phi$ that misleads the control network. We let $D(\{I_f[n]\}_f)$ denote the controller network, which takes as input a sequence of frames $\{I_f[n]\}$ in a single trajectory and generates a corresponding sequence of real-valued control signals, such as a steering angle at each time instant. Our goal is to find the value of $\Phi$ that maximizes deviation of these outputs from those for an unmodified environment. We cast this as minimization of a loss over our $T$ trajectories, i.e.,
\begin{equation}
 \label{eq:tloss}
  \Phi = \arg \min  - \sum_{t=1}^T \delta\left(D(\{I_f^t[n,\Phi]\}_f),D(\{\tilde{I}_f^t[n]\}_f)\right),
\end{equation}
in which $\delta(\cdot,\cdot)$ measures divergence between two sequences of control outputs.
In addition, we propose a physically meaningful variation of this, where we split the $T-1$ trajectories other than the one using actual control outputs into two subgroups: the one with positive and the other with negative perturbations, with both groups including the original trajectory.
Note that this is meaningful in the \emph{Groups} approach for generating trajectories above, but not for \emph{Random}.
After solving the two resulting problems independently, we can then choose the solution that has the highest divergence.
In our experiments where the control network outputs a sequence of steering angles, we use the absolute value of the sum of the differences between the angles as our divergence metric when we pool all trajectories together, and use the difference (for the positive subgroup) or negative difference (for the negative subgroup).
We do this because we would expect that positive perturbations will be more representative when we are trying to force steering further to the right, and negative perturbations are most physically meaningful when we aim to cause sharp steering to the left.

 We carry out either version of optimization iteratively using gradient descent with Adam~\citep{Adam} because, as shown next, we are able to compute gradients of the loss in \eqref{eq:tloss} with respect to the modification parameters $\Phi$. Since the controller network $D(\cdot)$ is assumed to be a neural network (or a differentiable function), we are able to use standard back-propagation to compute gradients $\nabla\left(I_f^t(n;\Phi)\right)$ of the loss with respect to each intensity in the images of the modified environment. The gradients with respect to the color parameters $\{x_k^C\}$ can then be computed based on \eqref{eq:composite} as:
 \begin{equation}
   \label{eq:cgrad}
   \nabla(x_k^C) = \sum_{t,f} \left(C_f^t\right)^T \left(\sum_{n} V_k(G_f^t(n); \Phi)~\nabla\left(I_f^t(n;\Phi)\right)\right).
 \end{equation}

 Computing gradients with respect to the shape parameters $\{x_k^S\}$ requires an approximation, since the mask functions $M(\cdot)$ are not generally differentiable with respect to these parameters. We adopt a simple local linearization approach: for every scalar parameter $\theta$ in the shape parameters $x_k^S$ for each figure, we discretize its range into a fixed set of equally separated values. Then, given the current (continuous) value of $\theta$, we let $\theta^-$ and $\theta^+$ represent the consecutive pair of values in this discrete set, such that $\theta^- \leq \theta \leq \theta^+$, and denote $\Phi_{\theta^-}$ and $\Phi_{\theta^+}$ the full set of current parameter values, with $\theta$ replaced by $\theta^+$ and $\theta^-$ respectively. We make the approximation that if $\alpha \in \mathbb{R}$ such that $\theta = \alpha \theta^+ + (1-\alpha) \theta^-$, then
 $
   I_f^t(n; \Phi) \approx \alpha I_f^t(n; \Phi_{\theta^+}) + (1-\alpha) I_f^t(n;\Phi_{\theta^-}).
 $
 Therefore, although we only use frames $I_f^t(n; \Phi)$ with the actual value of $\Phi$ as input to the control network, we also generate an extra pair of images $I_f^t(n; \Phi_{\theta^-}), I_f^t(n; \Phi_{\theta^+})$ for each frame for every element $\theta$ of the shape parameters. We then use these frames to compute parameter gradients as:

\begin{align}
   & \nabla(\alpha) = \sum_{t,f,n} \nabla\left(I_f^t(n;\Phi)\right)\left(I_f^t(n;\Phi_{\theta^+})-I_f^t(n;\Phi_{\theta^-})\right), \nonumber \\
   & \nabla(\theta) = \nabla(\alpha) \times (\theta^+-\theta^-)^{-1}.
\end{align}

\section{Experiments}
\label{sec:Experiment}
We experimentally demonstrate that our approach, which we refer to as \emph{GradOpt}, is both more efficient---requiring far fewer actual or simulated drives---and effective---in successfully finding attack patterns---than the state-of-the-art Bayesian Optimization (\emph{BO}) method~\citep{Boloor_2020_JSA}. 
To carry out a large scale evaluation, we perform our experiments in simulation using a packaged version of the Carla simulator~\citep{Dosovitskiy17} provided by~\citet{Boloor_2020_JSA} 
that allows the addition of painted road patterns to existing environments. 
Our experiments evaluate attacks against the neural network-based controller network that is included with Carla and uses only camera inputs and outputs steering angles; this network was trained using imitation learning. 
We run evaluations only on scenarios where this controller drives successfully without infractions in the unmodified environment.


Our experiments use 40 scenarios of
driving through a stretch of road in a virtual town. 
Each scenario begins an episode with the vehicle spawned at a given starting waypoint, and the controller is then tasked with reaching a defined destination waypoint. 
The episode runs until the vehicle reaches this destination or a time-limit expires (e.g., if the car crashes). Our scenarios are of three types: (a) the expected behavior is for the car to go \emph{straight} (16 scenarios), (b) veer \emph{left} (12 scenarios), or (c) \emph{right} (12 scenarios).
In each scenario, the attacker can draw a pattern on the road with the intent of causing the 
car to deviate from its intended path. 
We consider patterns that are unions of rectangles (i.e., each ``figure'' in Sec.~\ref{sec:physmap} is a rectangle), where the shape of each rectangle is determined by four parameters (i.e., $x_{k}^{C}\in\mathbb{R}^{4}$): rotation, width, length, and horizontal offset.\footnote{We center the rectangles vertically prior to rotation, as in~\citet{Boloor_2020_JSA}.}

We report results from optimizing 
shape and color parameters for different numbers of rectangles $K$, ranging from $K=1$ (7 parameters) to $K=5$ (35 parameters), and additionally for the single black rectangle case, also when optimizing only its shape  (4 parameters).
We learn these parameters with respect to the top-view co-ordinate frame of a canvas, that during evaluation will be passed to the simulator to be superimposed on the road (and then captured by the camera as part of the scene in all frames).

We train both BO and GradOpt in a simulated setting without any pedestrians or other cars.
We then evaluate the success of the attack on actual simulations with Carla, in terms of two metrics. The first measures deviation between the paths with and without the attack pattern without pedestrians or other cars. 
We define deviation as
\begin{equation}
  \label{eq:devdef}
  \mbox{Deviation} = \frac{1}{2T}\sum_{t=1}^{t=T} \min_{t'} |\widetilde{W}_t - W_{t'}|+\min_{t'} |W_t - \widetilde{W}_{t'}|,
\end{equation}
where $\widetilde{W}_{t}$ and $W_{t}$ are sequences of car locations when driving with and without the attack pattern, at a fixed set of time instances defined as those when the region of the road where the attack pattern would appear is visible.
Our second metric is the total infraction penalty when driving \emph{with} pedestrians and other cars, as defined by the Carla Autonomous Driving Challenge~\citep{CarlaChallenge} (for example, a lane violation carries a penalty of 2 points, 
hitting a static object or another vehicle of 6, 
hitting a pedestrian of 9, etc.). 
For each attack and scenario, we run 10 simulations, randomly spawning pedestrians and cars each time, and average the infraction penalty scores. 
Finally, while both BO and GradOpt are trained in a \emph{clear noon} weather setting, we measure infractions on that setting as well as three others: \emph{cloudy noon}, \emph{rainy noon}, and \emph{clear sunset}.

\subsection{Attack Optimization}

\noindent{\bf Proposed Method:} Our approach requires two steps for every scenario: (1) collecting a set of frame sequences and calibrating them, and (2) performing gradient-based optimization. For (1), we collect frames from $T=3$ trajectories; we compare different ways to generate these and the impact of varying $T$ below.
We estimate the homographies $G_{f}^{t}$ for every frame $f$ in trajectory $t$ by running additional simulations with calibration patterns painted on the canvas: we run 12 simulations for each trajectory, with 5 calibration points of different colors in each simulation, and use the 60 automatically detected correspondences to estimate the homography matrix for every frame. We also learn a common set of color transform parameters for all frames in all trajectories, which we obtain by running the simulator 22 times, each time with the entire canvas painted a different high-contrast color, on the unperturbed trajectory. In addition, we collect clean frames for each trajectory. 
Altogether, our method calls the simulator a total of \emph{61} times. 

Once we have a set of calibrated frames, we employ gradient-based optimization (see Sec.~\ref{sec:gopt}). 
We run the optimization with four different random starts, each for 250 iterations, for a total of 1000 iterations. We begin with a learning rate of 0.1 for each start, and drop it by a factor of 0.1 after the first 100, and the first 200 iterations. Each time we drop the learning rate, we also reset the parameters to those that yielded the lowest value of the loss thus far.

\noindent{\bf Bayesian Optimization:} We employ BO with the same objective as ours based on the same divergence metric, and closely follow the setting in \citet{Boloor_2020_JSA}---i.e., we run the optimization for a total of 1000 iterations, of which the first 400 are random exploration. Note that while this is the same number of iterations as we use for our method, every iteration of BO requires running a full episode with the Carla simulator.

\subsection{Results}
\noindent{\bf Run-time:} 
Recall that the training budget for both BO and GradOpt, is 1000 iterations. BO requires a simulator call at each iteration, with training times ranging  7-25 hours, depending on the scenario. In contrast, our method only calls the simulator 61 times for calibration. 
Ignoring the potential of parallelizing the iterations for the four random starts, GradOpt has a total running time of up to 2.5 hours per scenario, including both calibration and training. Thus, our method affords a significant advantage in computational cost and, as we show next, is also more successful at finding optimal attack patterns.

\begin{table*}[h]
\centering
\begin{tabular}{ccccccc}
\toprule
  & \multicolumn{6}{c}{\bf Deviation}\\
    K
  &\multicolumn{4}{c}{Group} & \multicolumn{1}{c}{Group-All} & \multicolumn{1}{c}{Random}\\%
  \cmidrule(lr){2-5}\cmidrule(lr){6-6} \cmidrule(lr){7-7}  
  (\#Rect.)
  &\multirow{1}{*}{T=1}
  &\multirow{1}{*}{T=3}
  &\multirow{1}{*}{T=5}
  &\multirow{1}{*}{T=7}
  &\multirow{1}{*}{T=3}
  &\multirow{1}{*}{T=3}\\
  \midrule
  1-b & \emph{0.89} & \emph{0.92} & \emph{1.03} & \emph{0.98} & \emph{0.92} & \emph{0.84}\\
  1   & \emph{0.86} & \emph{0.89} & \emph{0.89} & \emph{0.95} & \emph{0.94} & \emph{0.86}\\
  2   & \emph{1.05} & \emph{1.13} & \emph{1.07} & \emph{1.20} & \emph{1.13} & \emph{1.16}\\
  3   & \emph{1.11} & \emph{1.14} & \emph{1.22} & \emph{1.33} & \emph{1.26} & \emph{1.09}\\
  4   & \emph{1.21} & \emph{1.30} & \emph{1.28} & \emph{1.28} & \emph{1.32} & \emph{1.14}\\
  5   & \emph{1.23} & \emph{1.33} & \emph{1.33} & \emph{1.33} & \emph{1.33} & \emph{1.22}\\ \bottomrule
\end{tabular}
\begin{tabular}{ccccccc}
\toprule
  &\multicolumn{6}{c}{\bf Infraction Penalty} \\
    K&\multicolumn{4}{c}{Group}&
    \multicolumn{1}{c}{Group-All}& \multicolumn{1}{c}{Random} \\
  \cmidrule(lr){2-5}\cmidrule(lr){6-6} \cmidrule(lr){7-7}
  (\#Rect.)
  &\multirow{1}{*}{T=1}
  &\multirow{1}{*}{T=3}
  &\multirow{1}{*}{T=5}
  &\multirow{1}{*}{T=7}
  &\multirow{1}{*}{T=3}
  &\multirow{1}{*}{T=3}\\
  \midrule
  1-b &  \emph{3.80} &\emph{4.25} &\emph{4.19} & \emph{4.43} & \emph{3.82} & \emph{3.98}\\
  1   &  \emph{4.18} &\emph{5.20} &\emph{5.54} & \emph{4.97} & \emph{4.73} & \emph{4.68}\\
  2   &  \emph{3.86} &\emph{5.16} &\emph{5.33} & \emph{5.85} & \emph{4.96} & \emph{4.88}\\
  3   &  \emph{3.79} &\emph{6.04} &\emph{5.62} & \emph{6.50} & \emph{5.28} & \emph{4.97}\\
  4   &  \emph{5.04} &\emph{6.35} &\emph{5.45} & \emph{6.00} & \emph{5.39} & \emph{5.39}\\
  5   &  \emph{4.17} &\emph{6.65} &\emph{6.29} & \emph{6.26} & \emph{6.54} & \emph{5.83}\\ \bottomrule
\end{tabular}
\caption{Ablation analysis of variations of GradOpt.}
\label{T:ablation}
\end{table*}

\noindent{\bf Ablation Analysis of GradOpt:} 
First we identify the best variation of GradOpt, in terms of the choice of $T$, the choice between \emph{Random} (randomly perturbing each trajectory) and \emph{Group} (generating pairs of perturbed trajectories, one with positive and another with negative perturbations), and for the latter, whether we pool all trajectories in one optimization problem (\emph{Group-All}) or separately optimize only positively/negatively perturbed trajectories, respectively (\emph{Group}).
The results in Table~\ref{T:ablation} show that \emph{Group} has the best performance, particularly in terms of infraction penalties.
Moreover, $T=3$ yields significant improvement over $T=1$, but further increasing $T$ does not.
This shows that our approach that makes use of perturbed trajectories to counter the car's self-correcting behavior is indeed important, and remarkably effective, requiring only 2 perturbed trajectories.
Moreover, we can see that separately solving the problem with only positive, and only negative, perturbation (both including the baseline), rather than pooling these into a single objective, is important in yielding more infractions, even though there is no difference in terms of divergence.
The intuition for this is that pooling only, say, positively perturbed trajectories makes them consistent with the goal of the optimization (which is to steer sharply to the right, in that case), and the attack is better able to counter self-correcting behavior of the vehicle.
Thus, we use the \emph{Group} variant of GradOpt in the sequel.

\begin{table*}[h]
\centering
\begin{tabular}{cp{10pt}cp{10pt}p{0pt}p{10pt}cp{10pt}}
\toprule
  & \multicolumn{3}{c}{\bf Deviation} & & \multicolumn{3}{c}{\bf ~Infraction Penalty} \\
  K
  &\multirow{2}{*}{ BO}&\multirow{2}{*}{GradOpt}&\multirow{2}{*}{\% $\geq$}& 
  &\multirow{2}{*}{ BO}&\multirow{2}{*}{GradOpt}&\multirow{2}{*}{\% $\geq$}\\
  (\#Rect.)&&&&\\\midrule
  1-b & 0.85 & \emph{0.92} & \emph{53\%} && 3.85 & \emph{4.25} &\emph{80\%}\\
  1   & 0.85 & \emph{0.89} & \emph{45\%} && 4.30 & \emph{5.20} &\emph{69\%}\\
  2   & 0.79 & \emph{1.13} & \emph{70\%} && 4.55 & \emph{5.16} &\emph{76\%}\\
  3   & 0.93 & \emph{1.14} & \emph{78\%} && 4.53 & \emph{6.04} &\emph{74\%} \\
  4   & 0.70 & \emph{1.30} & \emph{90\%} && 3.79 & \emph{6.35} &\emph{81\%}\\
  5   & 0.84 & \emph{1.33} & \emph{82\%} && 4.73 & \emph{6.65} &\emph{80\%}\\ \bottomrule
\end{tabular}
\caption{Average deviation 
and infraction penalties 
over all scenarios for GradOpt and BO, when optimizing parameters of different numbers of rectangles (1-b optimizes only the shape of one black rectangle) in 
``clear noon'' weather.
The $\%\ge$ column reports the percentage of instances where GradOpt has $\ge$ score than BO. 
}
\label{T:base-all}
\end{table*}

\begin{figure}[h]
  \centering
  \includegraphics[width=0.82\columnwidth]{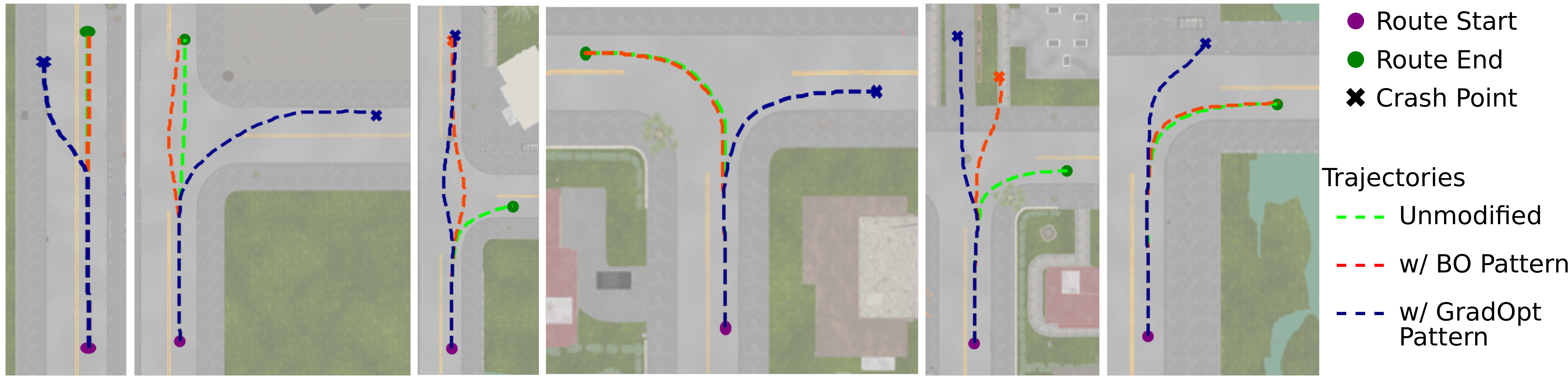}
  \caption{Trajectory deviations induced by GradOpt and BO for 6 example scenarios. 
  }
\label{fig:traj}
\end{figure}

\noindent{\bf Efficacy of GradOpt:} 
Table~\ref{T:base-all} compares GradOpt and BO
in terms of the metrics deviation and infraction penalty discussed above, computed with simulations in the same standard weather setting as used for attack optimization. 
We report the averages, as well as the percentage of scenarios when GradOpt outperforms BO as comparison statistics.
We find that GradOpt is significantly more successful than BO, despite also being computationally less expensive as discussed earlier. It has higher average values of both deviation and infraction penalties, with the gap growing significantly for higher values of $K$---indicating that GradOpt is much better able to carry out optimization in a higher-dimensional parameter space and leverage the ability to use more complex patterns. Moreover, we find that it yields attack patterns that are as or more successful than BO in more than 70\% of cases in all settings, with the advantage rising to 82\% for $K=5$. Figure \ref{fig:traj} shows some example scenarios comparing trajectory deviations induced by the attack patterns discovered by the two algorithms. 
Additional illustrations and visualizations are provided in the Supplement.


We take a closer look at the vulnerability of different types of scenarios by separately reporting infraction penalties for each in Table \ref{T:pertype}. In addition, we also report the corresponding infraction penalties computed in simulations \emph{without} cars or pedestrians in Table~\ref{T:pertypeScoreWithoutVP}.
We see that scenarios where the expected behavior is driving straight are the hardest to attack, likely because they are the simplest to drive in. BO tends to achieve only a moderate infraction score in these settings, even at higher values of $K$. In contrast, GradOpt reveals that even these scenarios are in fact vulnerable when one is allowed to consider more complex adversarial patterns---achieving an average infraction penalty that is 
significantly
higher than BO at $K=5$. Conversely, driving right scenarios are the most vulnerable with both methods being successful even with simple patterns, with GradOpt again yielding higher deviation and more infractions.

\begin{table*}[h]
\centering
\begin{tabular}{cccccccccccc}
\toprule
  K& \multicolumn{3}{c}{\bf Straight} &~~& \multicolumn{3}{c}{\bf Left} &~~& \multicolumn{3}{c}{\bf Right}\\
  \#Rect.& {BO}&{GradOpt}&{\% $\geq$}&& {BO}&{GradOpt}&{\% $\geq$}&& {BO}&{GradOpt}&{\% $\geq$}\\\midrule
  1-b & 2.24 & \emph{2.33} & \emph{84\%}  && 4.70 & \emph{4.15} & \emph{76\%} && 5.14 & \emph{6.90} & \emph{78\%}\\
  1   & 2.02 & \emph{2.81} & \emph{77\%}  && 5.53 & \emph{4.79} & \emph{60\%} && 6.11 & \emph{8.78} & \emph{68\%}\\
  2   & 1.91 & \emph{2.73} & \emph{91\%} && 5.28 & \emph{5.28} & \emph{63\%} && 7.33 & \emph{8.28} & \emph{68\%}\\
  3   & 2.74 & \emph{4.30} & \emph{84\%}  && 5.47 & \emph{5.45} & \emph{67\%} && 5.98 & \emph{8.94} & \emph{70\%}\\
  4   & 0.97 & \emph{5.16} & \emph{95\%} && 5.00 & \emph{4.55} & \emph{72\%} && 6.33 & \emph{9.73} & \emph{73\%}\\
  5   & 1.69 & \emph{4.86} & \emph{86\%}  && 5.60 & \emph{6.65} & \emph{76\%} && 7.92 & \emph{9.04} & \emph{76\%}\\
  \bottomrule
\end{tabular}
\caption{Infraction penalties by scenario type (driving straight, left, or right) in  ``clear noon'' conditions.
}
\label{T:pertype}
\end{table*}

\begin{table*}[h]
\centering
\begin{tabular}{cccccccccccc}
\toprule
  K& \multicolumn{3}{c}{\bf Straight} &~~& \multicolumn{3}{c}{\bf Left} &~~& \multicolumn{3}{c}{\bf Right}\\
  \#Rect.& {BO}&{GradOpt}&{\% $\geq$}&& {BO}&{GradOpt}&{\% $\geq$}&& {BO}&{GradOpt}&{\% $\geq$}\\\midrule
  1-b & 2.25 & \emph{1.50} & \emph{62\%}  && 1.83 & \emph{2.67} & \emph{67\%} && 3.00 & \emph{6.50} & \emph{92\%}\\
  1   & 1.38 & \emph{1.88} & \emph{69\%}  && 3.17 & \emph{2.33} & \emph{67\%} && 5.33 & \emph{7.00} & \emph{67\%}\\
  2   & 0.62 & \emph{1.62} & \emph{94\%} && 2.33 & \emph{2.17} & \emph{67\%} && 5.50 & \emph{8.33} & \emph{83\%}\\
  3   & 2.38 & \emph{2.62} & \emph{75\%}  && 3.83 & \emph{3.83} & \emph{75\%} && 5.50 & \emph{7.33} & \emph{75\%}\\
  4   & 0.50 & \emph{2.25} & \emph{100\%} && 2.17 & \emph{1.83} & \emph{83\%} && 4.67 & \emph{7.00} & \emph{92\%}\\
  5   & 1.75 & \emph{3.00} & \emph{81\%}  && 3.17 & \emph{4.17} & \emph{83\%} && 5.67 & \emph{7.00} & \emph{67\%}\\\bottomrule
\end{tabular}
\caption{Infraction penalties \emph{without cars or pedestrians},  i.e., infraction penalties computed with only \emph{static objects}, in standard ``clear noon'' simulations for each type of scenario.}
\label{T:pertypeScoreWithoutVP}
\end{table*}

\begin{table*}[h]
\centering
\begin{tabular}{cccccccccccc}
\toprule
  K& \multicolumn{3}{c}{\bf Cloudy Noon} &~~& \multicolumn{3}{c}{\bf Rainy Noon} &~~& \multicolumn{3}{c}{\bf Clear Sunset}\\
  \#Rect.& {BO}&{GradOpt}&{\% $\geq$}&& {BO}&{GradOpt}&{\% $\geq$}&& {BO}&{GradOpt}&{\% $\geq$}\\\midrule
  1-b & 2.29 & \emph{3.56} & \emph{78\%} && 2.69 & \emph{3.14} & \emph{83\%} && 2.41 & \emph{3.19} & \emph{89\%}\\
  1   & 3.19 & \emph{3.61} & \emph{82\%} && 2.98 & \emph{4.17} & \emph{86\%} && 3.36 & \emph{3.12} & \emph{85\%}\\
  2   & 2.87 & \emph{4.45} & \emph{86\%} && 4.03 & \emph{4.45} & \emph{78\%} && 2.92 & \emph{3.85} & \emph{84\%}\\
  3   & 3.06 & \emph{5.21} & \emph{85\%} && 3.28 & \emph{5.40} & \emph{82\%} && 2.60 & \emph{4.38} & \emph{87\%}\\
  4   & 2.39 & \emph{4.96} & \emph{89\%} && 3.18 & \emph{4.88} & \emph{85\%} && 2.38 & \emph{3.68} & \emph{84\%}\\
  5   & 3.56 & \emph{5.87} & \emph{83\%} && 3.81 & \emph{5.26} & \emph{77\%} && 2.67 & \emph{4.37} & \emph{83\%}\\\bottomrule
\end{tabular}
\caption{Infraction penalties 
over all scenarios 
with weather conditions different from that used for optimizing attacks (``clear noon'').}
\label{T:climate}
\end{table*}

\noindent{\bf Transferability:}
We evaluate the robustness of our adversarial test generation approach by evaluating the success of generated adversarial perturbations in different climate and visibility conditions than those used for attack optimization. Table~\ref{T:climate} presents results for simulations with four such climate settings, and as expected, we find that both BO and GradOpt do see a drop in penalty scores compared to the standard setting in Table~\ref{T:base-all}. Nevertheless, most of the attacks induce infractions, especially at higher values of $K$, with GradOpt again being significantly more successful than BO.

\section{Conclusion and Ethical Considerations}
\label{sec:ConclusionEthics}
A great deal of attention has been devoted to understanding adversarial perturbations in computational perception, with autonomous driving the most common motivation. However, most prior research has considered these to be independent for each input image. In contrast, autonomous driving is dynamic, and even if a perturbation succeeds to fool the car in a particular frame, it can fail in another frame, and the car can self-correct. Thus, to fully understand vulnerabilities in autonomous vehicle architectures we need to evaluate them in an end-to-end fashion using driving scenarios. However, this is inherently challenging because the resulting experimentation, be it using a simulator or actual driving, is extraordinarily time consuming. We developed a novel framework that allows us to largely avoid costly driving experiments, relying instead on a novel compositing approach which is fully differentiable. Our approach is significantly more potent at discovering physically realizable adversarial examples than prior art, while also requiring far fewer runs of actual driving trials or simulations. Moreover, we find that the vulnerabilities we discover are robust, and persist even under variability in weather conditions.

Security research in general, and security of AI in particular, inevitably raises ethical considerations when the primary contribution develops novel attacks to demonstrate vulnerabilities in systems.
We, too, have developed a novel framework that is able to exploit autonomous driving systems far more effectively than the state of the art.
Moreover, our focus on \emph{physically realizable} adversarial examples (that is, examples that are designed to modify objects in the world---road surface, in our case---rather than digital images at the level of pixels) brings our work even closer to reality than many other efforts that attack perception by adding adversarial noise at the level of pixels.
This line of research, however, is absolutely critical: if we are to put autonomous cars on the roads without adequately stress-testing them, the consequences of failures can be catastrophic.

Our approach targets primarily simulation-based stress-testing of autonomous vehicles.
We do this for two reasons.
First, the fact that perturbations are restricted to simulations means that our approach cannot be used "out-of-the-box" to directly cause accidents.
Second, simulations, coupled with our method, enable far highly scalable testing of vehicles, identifying vulnerabilities before the autonomous vehicle architecture is deployed on actual roads, thereby reducing the likelihood of unanticipated vulnerabilities manifesting themselves when it truly matters.


\bibliography{references}
\bibliographystyle{plainnat}

\clearpage
\appendix
\section*{Supplement}
\renewcommand{\subsection}{\section}
In this supplement, we provide further details of our shape parameterization and calibration, as well as additional results and visualizations from our experiments.

\subsection{Pattern Shape Parameters}

Each rectangle in our shape is parameterized by four values $x_{k}^{S}=[w,l,o,\theta]$, corresponding to width, length, horizontal offset, and rotation or orientation. These parameters are defined with respect to the top-down view of a $400\times 400$ pixel ``canvas'' that is composited onto the road surface. Each rectangle is first drawn aligned with the $x-$ and $y-$axes of this canvas to be of width $w$ and length $l$ pixels, and vertically centered and horizontally offset so that its left edge is at $x=o$, and then rotated about the center of the canvas by angle $\theta$. Portions of rectangles that lay outside the canvas after this process were clipped from the pattern. Our parameterization expands on the one originally used by \citet{Boloor_2020_JSA} in two respects: by searching over length $l$ instead of fixing it to the length of the canvas, and having a separate orientation $\theta$ for each rectangle rather than a common one for all rectangles.

\subsection{Calibration Details}

We estimate homographies between the canvas and each frame from 60 corresponding pairs as described in Sec.~\ref{sec:Experiment}, using a direct linear transform. While doing so, we ensure the matrix has the correct sign so that homogeneous co-ordinates of points projected in the frame have a positive third co-ordinate when they are visible, and a negative one when they are ``behind the camera''. When compositing patterns on the frame, this allows us to retain only the portion of the pattern that would be visible from that viewpoint. The color transforms are estimated simply from the color correspondences using a least-squares fit.

\begin{table}[h]
\centering
\begin{tabular}{cp{10pt}cp{10pt}p{0pt}p{10pt}cp{10pt}}
\toprule
  & \multicolumn{3}{c}{\bf Deviation} & & \multicolumn{3}{c}{\bf ~Infraction Penalty} \\
  K
  &\multirow{2}{*}{BO}&\multirow{2}{*}{GradOpt}&\multirow{2}{*}{\% $\geq$}&~~~~~~
  &\multirow{2}{*}{BO}&\multirow{2}{*}{GradOpt}&\multirow{2}{*}{\% $\geq$}\\
  \#Rect.&&&&\\\midrule
  1-b & 0.97 & \emph{1.03} & \emph{45\%} && 1.45 & \emph{1.95} &\emph{80\%}\\
  1   & 0.49 & \emph{0.94} & \emph{72\%} && 1.90 & \emph{2.25} &\emph{82\%}\\
  2   & 0.55 & \emph{1.08} & \emph{82\%} && 1.80 & \emph{3.10} &\emph{90\%}\\
  3   & 0.50 & \emph{1.11} & \emph{88\%} && 1.90 & \emph{3.40} &\emph{90\%} \\
  4   & 0.57 & \emph{1.11} & \emph{90\%} && 1.90 & \emph{3.00} &\emph{85\%}\\
  5   & 0.51 & \emph{1.19} & \emph{88\%} && 2.55 & \emph{3.15} &\emph{78\%}\\ \bottomrule
\end{tabular}
\caption{Deviation and infraction penalties, \emph{both} computed without cars or pedestrians, over all scenarios in ``cloudy noon'' conditions.}
\label{T:base-cloudy-all}
\end{table}

\begin{table}[h]
\centering
\begin{tabular}{cp{10pt}cp{10pt}p{0pt}p{10pt}cp{10pt}}
\toprule
  & \multicolumn{3}{c}{\bf Deviation} & & \multicolumn{3}{c}{\bf ~Infraction Penalty} \\
  K
  &\multirow{2}{*}{BO}&\multirow{2}{*}{GradOpt}&\multirow{2}{*}{\% $\geq$}&~~~~~~
  &\multirow{2}{*}{BO}&\multirow{2}{*}{GradOpt}&\multirow{2}{*}{\% $\geq$}\\
  \#Rect.&&&&\\\midrule
  1-b & 0.72 & \emph{1.02} & \emph{57\%} && 1.15 & \emph{2.30} &\emph{88\%}\\
  1   & 0.50 & \emph{0.86} & \emph{72\%} && 1.45 & \emph{2.45} &\emph{90\%}\\
  2   & 0.51 & \emph{0.88} & \emph{75\%} && 1.95 & \emph{2.85} &\emph{82\%}\\
  3   & 0.57 & \emph{1.10} & \emph{75\%} && 1.90 & \emph{3.90} &\emph{85\%} \\
  4   & 0.58 & \emph{0.97} & \emph{80\%} && 1.75 & \emph{2.70} &\emph{85\%}\\
  5   & 0.58 & \emph{1.24} & \emph{90\%} && 2.50 & \emph{3.80} &\emph{82\%}\\ \bottomrule
\end{tabular}
\caption{Deviation and infraction penalties, \emph{both} computed without cars or pedestrians, over all scenarios in ``rainy noon'' conditions.}
\label{T:base-Rain-all}
\end{table}

\begin{table}[h]
\centering
\begin{tabular}{cp{10pt}cp{10pt}p{0pt}p{10pt}cp{10pt}}
\toprule
  & \multicolumn{3}{c}{\bf Deviation} & & \multicolumn{3}{c}{\bf ~Infraction Penalty} \\
  K
  &\multirow{2}{*}{BO}&\multirow{2}{*}{GradOpt}&\multirow{2}{*}{\% $\geq$}& 
  &\multirow{2}{*}{BO}&\multirow{2}{*}{GradOpt}&\multirow{2}{*}{\% $\geq$}\\
  \#Rect.&&&&\\\midrule
  1-b & 0.49 & \emph{0.69} & \emph{57\%} && 1.15 & \emph{1.30} &\emph{88\%}\\
  1   & 0.40 & \emph{0.62} & \emph{62\%} && 1.75 & \emph{2.00} &\emph{82\%}\\
  2   & 0.29 & \emph{0.71} & \emph{78\%} && 1.30 & \emph{1.90} &\emph{85\%}\\
  3   & 0.38 & \emph{0.96} & \emph{90\%} && 0.95 & \emph{3.30} &\emph{95\%} \\
  4   & 0.35 & \emph{1.07} & \emph{85\%} && 1.60 & \emph{2.25} &\emph{85\%}\\
  5   & 0.44 & \emph{1.02} & \emph{85\%} && 1.90 & \emph{2.75} &\emph{85\%}\\ \bottomrule
\end{tabular}
\caption{Deviation and infraction penalties, \emph{both} computed without cars or pedestrians, over all scenarios in ``clear sunset'' conditions.}
\label{T:base-sunset-all}
\end{table}


\subsection{Additional Results}

In our main evaluation, we reported deviations (without cars or pedestrian) only in the overall evaluation in Table~\ref{T:base-all} and reported infraction penalties primarily \emph{with} cars and pedestrians for all other comparisons. 
For completeness, we report deviation scores for those comparisons here, as well as infraction penalties computed in simulations without cars or pedestrians (with the caveat that some of the highest penalties defined by the challenge are for collisions with pedestrians and cars).

Tables~\ref{T:base-cloudy-all}-\ref{T:base-sunset-all} report both deviation and car and pedestrian-free infraction penalty scores for simulations in the different non-standard weather conditions. Table~\ref{T:pertypeDeviation} reports deviation scores separately for different types of scenarios. We find that these results are qualitatively consistent with those in our main evaluation in Sec.~\ref{sec:Experiment}.

\begin{table*}[!b]
\centering
\begin{tabular}{cccccccccccc}
\toprule
  K& \multicolumn{3}{c}{\bf Straight} &~~& \multicolumn{3}{c}{\bf Left} &~~& \multicolumn{3}{c}{\bf Right}\\
  \#Rect.& {BO}&{GradOpt}&{\% $\geq$}&& {BO}&{GradOpt}&{\% $\geq$}&& {BO}&{GradOpt}&{\% $\geq$}\\\midrule
  1-b & 0.72 & \emph{0.52} & \emph{38\%}  && 0.89 & \emph{1.25} & \emph{58\%} && 0.99 & \emph{1.12} & \emph{67\%}\\
  1   & 0.49 & \emph{0.50} & \emph{31\%}  && 0.89 & \emph{1.23} & \emph{75\%} && 1.30 & \emph{1.06} & \emph{33\%}\\
  2   & 0.47 & \emph{0.70} & \emph{69\%} && 0.94 & \emph{1.47} & \emph{75\%} && 1.05 & \emph{1.35} & \emph{67\%}\\
  3   & 0.62 & \emph{0.79} & \emph{81\%}  && 1.23 & \emph{1.35} & \emph{67\%} && 1.02 & \emph{1.39} & \emph{83\%}\\
  4   & 0.40 & \emph{0.87} & \emph{94\%} && 0.93 & \emph{1.75} & \emph{92\%} && 0.86 & \emph{1.42} & \emph{83\%}\\
  5   & 0.54 & \emph{0.93} & \emph{81\%}  && 0.94 & \emph{1.83} & \emph{92\%} && 1.15 & \emph{1.36} & \emph{75\%}\\\bottomrule
\end{tabular}
\caption{Deviations in simulations without cars or pedestrians, in standard ``clear noon'' weather conditions, for each type of scenario. }
\label{T:pertypeDeviation}
\end{table*}


\begin{figure*}[t]
  \centering
  \includegraphics[width=0.5\columnwidth]{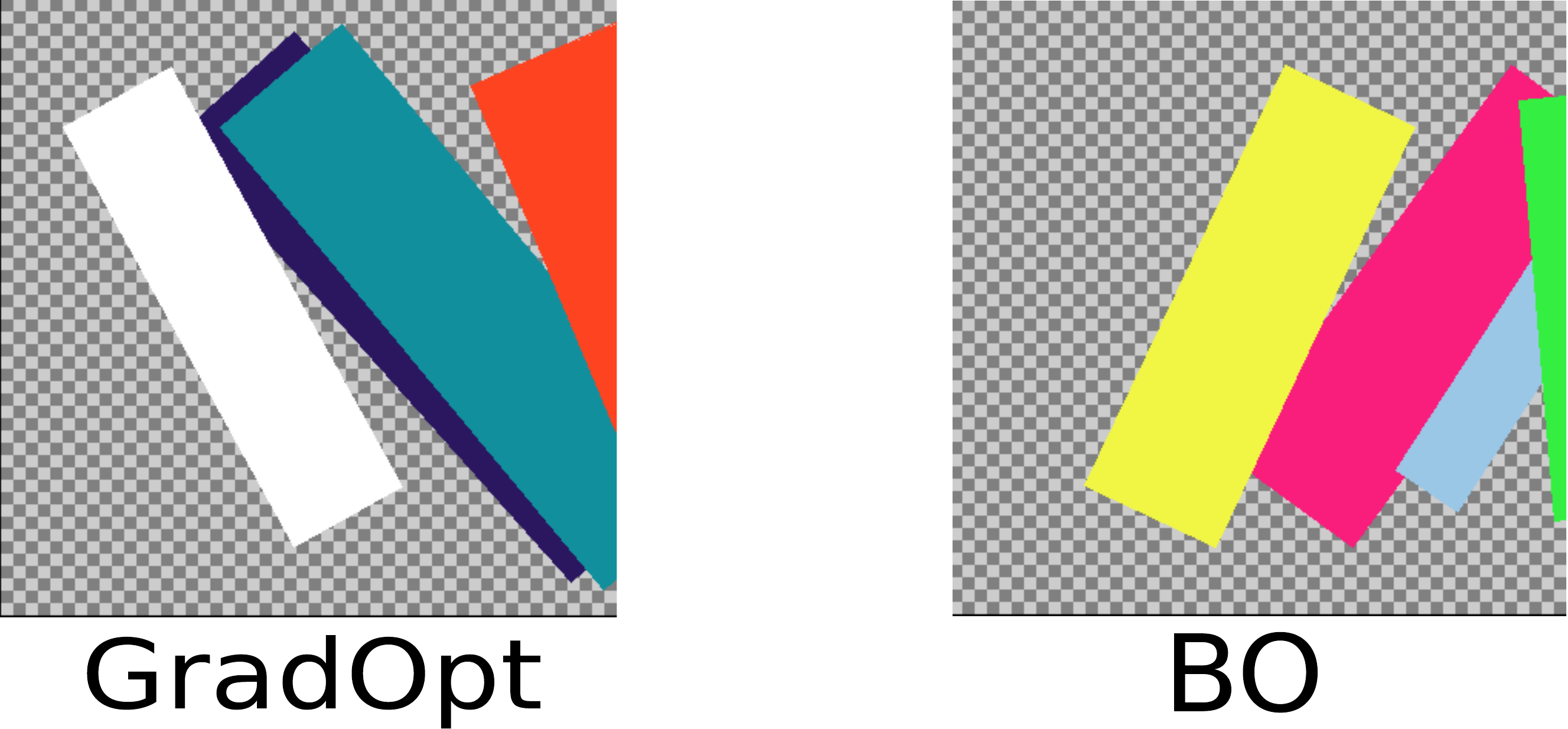}
  \caption{Attack patterns with $K=4$ rectangles (with transparent background) returned by GradOpt and BO for the example ``drive straight'' scenario illustrated in Figures \ref{fig:visual-gradOptReal} and \ref{fig:visual-gradOptBase}.}
\label{fig:perturbations}
\end{figure*}

\subsection{Visualization}

\begin{figure*}[h]
  \centering
  \textbf{Simulations with GradOpt Pattern}\\
  \includegraphics[width=0.95\columnwidth]{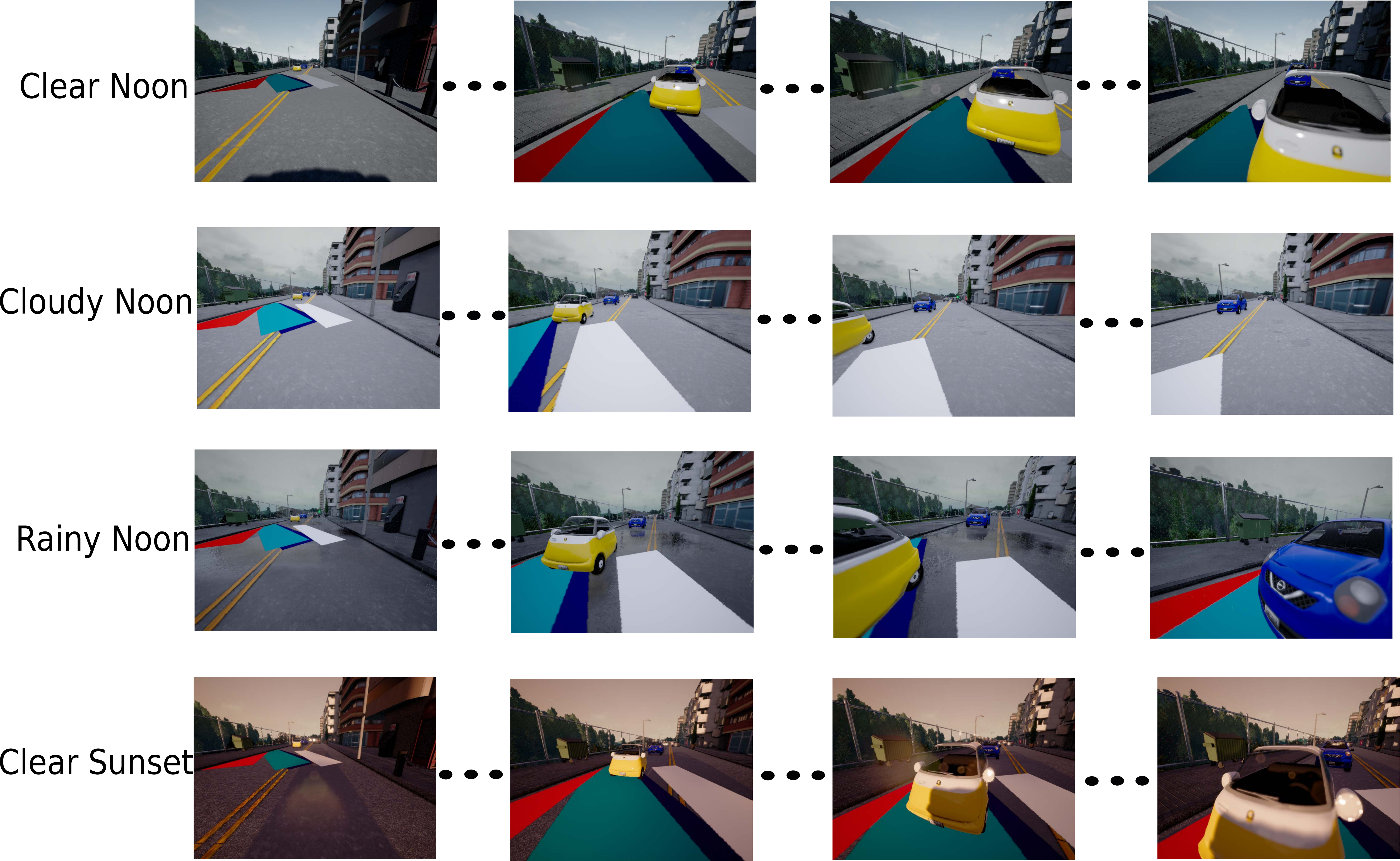}\\~\\
  \textbf{Simulations with BO Pattern}\\
  \includegraphics[width=0.95\columnwidth]{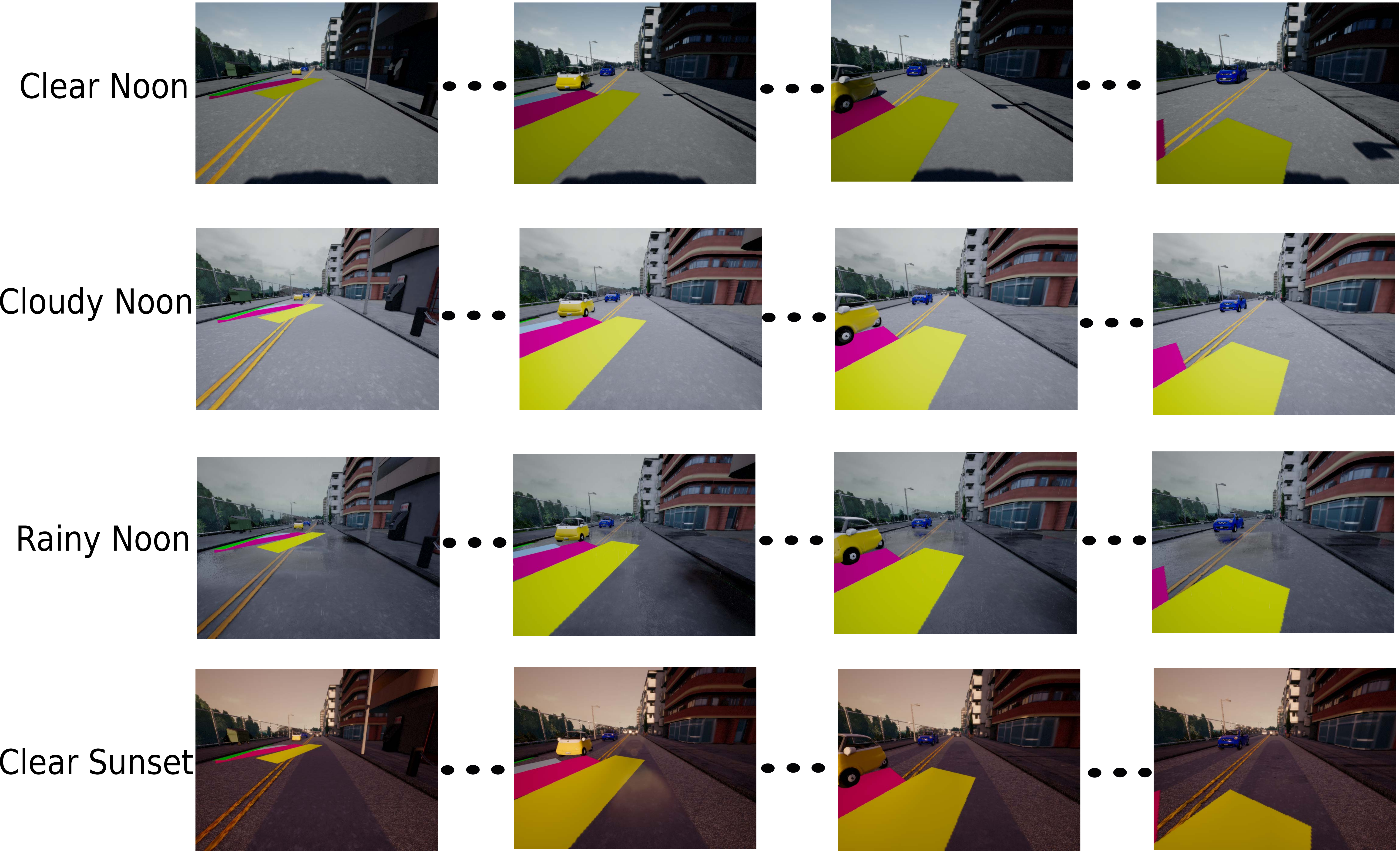}\\
  \caption{Frames from driving simulations, with cars and pedestrians in different weather conditions, after introducing attack patterns from GradOpt (top) and BO (bottom).}
\label{fig:visual-gradOptReal}
\end{figure*}

\begin{figure*}[h]
  \centering
  \textbf{Simulations with GradOpt Pattern}\\
  \includegraphics[width=0.95\columnwidth]{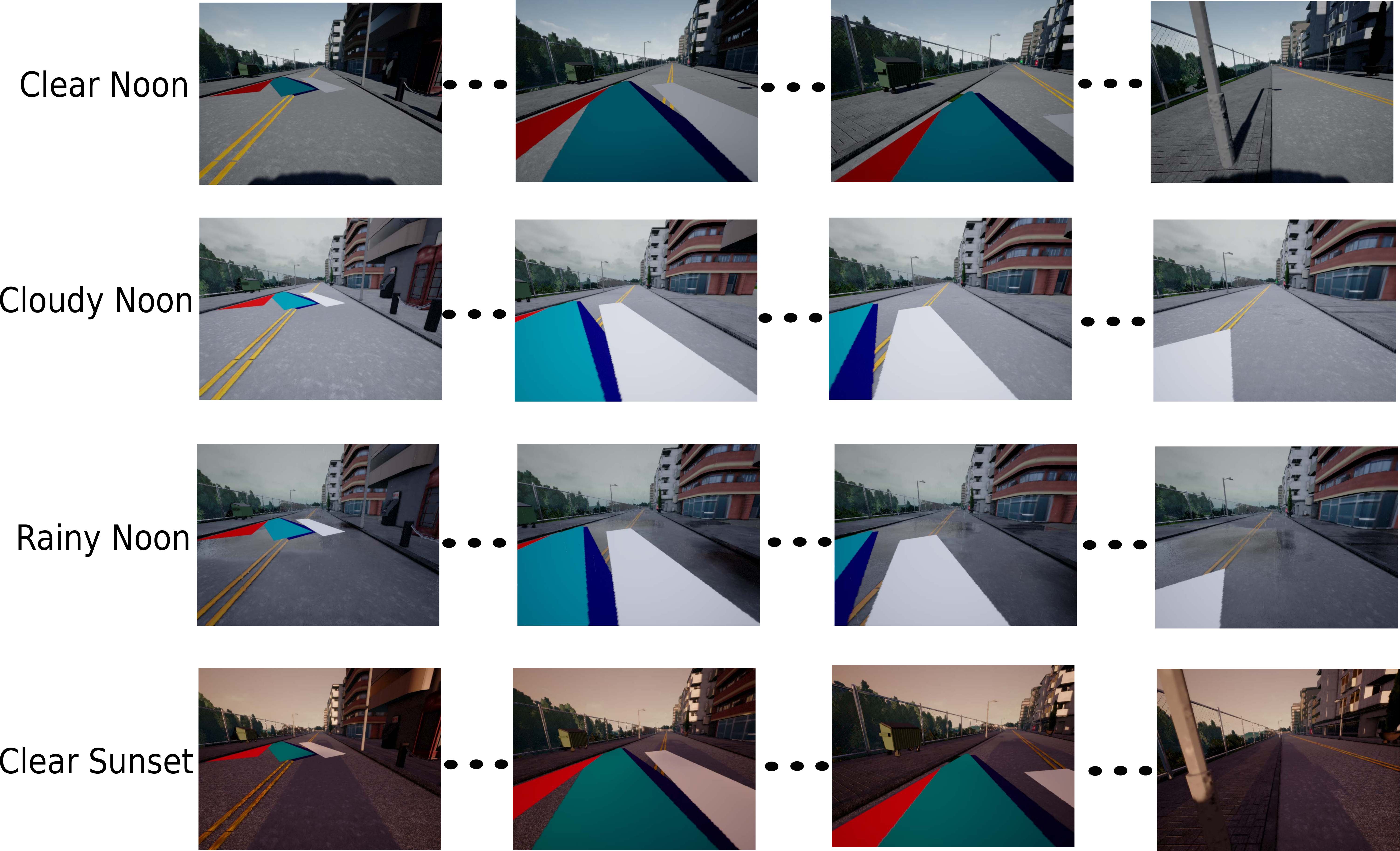}\\~\\
  \textbf{Simulations with BO Pattern}\\
  \includegraphics[width=0.95\columnwidth]{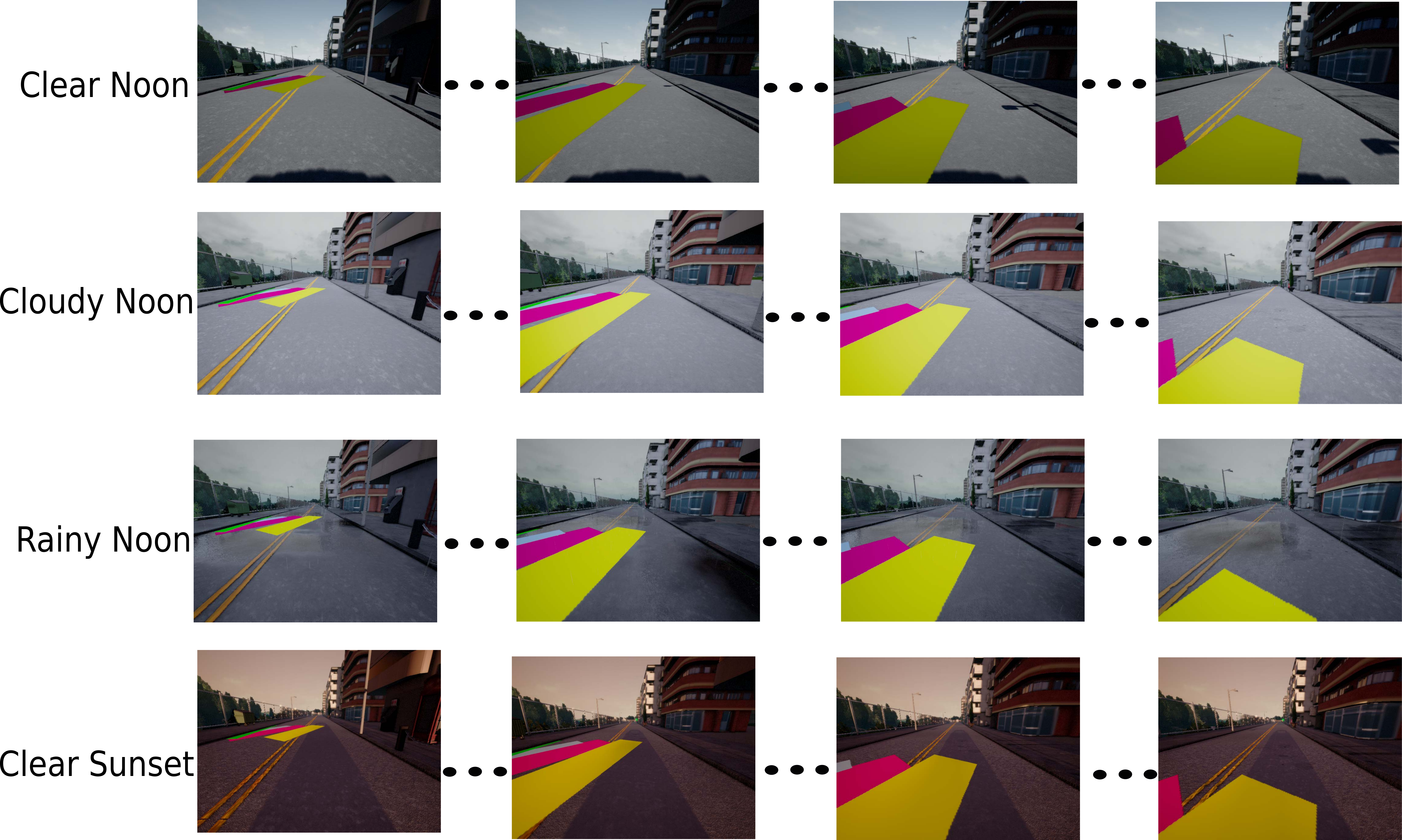}\\
  \caption{Frames from driving simulations, without cars or pedestrians in different weather conditions, after introducing attack patterns from GradOpt (top) and BO (bottom).}
\label{fig:visual-gradOptBase}
\end{figure*}

Finally, we use an example ``drive straight'' scenario to visualize the behavior of the controller when driving with attack patterns, with $K=4$ rectangles each, returned by GradOpt and BO. We show these patterns, in the top-down canvas view, in Fig.~\ref{fig:perturbations}. Then, we show frames from the vehicle's camera feed as it drives through the road with the respective patterns painted on the road in various climate conditions, in simulations with pedestrians and cars in Fig.~\ref{fig:visual-gradOptReal}, and without in Fig.~\ref{fig:visual-gradOptBase}. For this scenario, the pattern returned by BO is unable to cause a significant deviation in the vehicle's trajectory as it drives across the stretch of road with the pattern painted on it. In contrast, GradOpt's pattern is able to cause the car to veer sharply to the left in all but the ``cloudy noon'' climate setting---causing it to crash into another car in Fig.~\ref{fig:visual-gradOptReal} and veer sharply to the left in "clear noon" and "clear sunset" climate settings into the opposite sidewalk in Fig.~\ref{fig:visual-gradOptBase}.

\end{document}